# Distortions in Judged Spatial Relations in Large Language Models[1]

Nir Fulman[a,‡], Abdulkadir Memduhoğlu[a,b,‡], Alexander Zipf[a,c]

nir.fulman@uni-heidelberg.de; memduhoglu@uni-heidelberg.de; zipf@uni-heidelberg.de

[a] GIScience Chair, Institute of Geography, Heidelberg University, Heidelberg, Germany
[b] Department of Geomatic Engineering, Faculty of Engineering, Harran University, Sanliurfa, Türkiye
[c] HeiGIT at Heidelberg University, Heidelberg, Germany

We present a benchmark for assessing the capability of Large Language Models (LLMs) to discern intercardinal directions between geographic locations and apply it to three prominent LLMs: GPT-3.5, GPT-4, and Llama-2. This benchmark specifically evaluates whether LLMs exhibit a hierarchical spatial bias similar to humans, where judgments about individual locations' spatial relationships are influenced by the perceived relationships of the larger groups that contain them. To investigate this, we formulated 14 questions focusing on well-known American cities. Seven questions were designed to challenge the LLMs with scenarios potentially influenced by the orientation of larger geographical units, such as states or countries, while the remaining seven targeted locations were less susceptible to such hierarchical categorization. Among the tested models, GPT-4 exhibited superior performance with 55 percent accuracy, followed by GPT-3.5 at 47 percent, and Llama-2 at 45 percent. The models showed significantly reduced accuracy on tasks with suspected hierarchical bias. For example, GPT-4's accuracy dropped to 33 percent on these tasks, compared to 86 percent on others. However, the models identified the nearest cardinal direction in most cases, reflecting their associative learning mechanism, thereby embodying human-like misconceptions. We discuss avenues for improving the spatial reasoning capabilities of LLMs.

---

[1] This manuscript has been accepted for publication in *The Professional Geographer*.
[‡] These authors contributed equally to this work.

# Introduction

The technological landscape in artificial intelligence has been significantly shaped by the advancement of large language models (LLMs) such as the Generative Pre-trained Transformer (GPT) series (OpenAI 2023), along with others like Llama-2 (Touvron et al. 2023). These models have demonstrated remarkable abilities in understanding and generating text that resembles human writing across a broad range of tasks, including writing code, solving mathematical problems, and logical reasoning, demonstrating their broad applicability across different fields (Bubeck et al. 2023; Yuan et al. 2023; Liu et al. 2023).

Attention has also been directed towards the utility of LLMs within the geographic data analysis domain. Despite facing challenges with numerical accuracy and the inference of abstract relationships (Li and Ning, 2023; Cohn, 2023), these models have demonstrated effectiveness in real-world applications that leverage their extensive database of geographic information, complemented by inference capabilities. Examples of such applications include recalling populations of countries, estimating distances between cities, and planning itineraries in real-world geographical settings (Roberts et al., 2023). The practical utility of LLMs in processing spatial data and reasoning supports the case for their continued exploration and development in this direction.

Furthermore, recent research has unveiled non-spatial biases in LLMs, including logical and cognitive distortions (Gallegos et al., 2023). Cognitive psychology has long recognized systematic errors in human mapping memory, showing how our spatial perceptions often deviate from actual geography (Tversky, 1992). Since the data LLMs are trained on may include human errors and oversimplification of geographical details in such texts, and given these models' tendency to form conceptual associations favoring narrative coherence over geographical fidelity (Vaswani et al. 2017), we hypothesize that LLMs might replicate these human-like spatial biases. Specifically, we ask: "Do LLMs exhibit hierarchy bias in their spatial reasoning?" We examine this possibility by introducing a benchmark aimed at evaluating LLMs' ability to determine intercardinal directions between cities, with a focus on identifying potential hierarchy biases similar to those observed in human cognitive mapping processes.

# Background

*LLM utilization in geographic data analysis*

Investigations into the utility of LLMs within geography have aimed to assess their potential uses and identify the limitations they encounter. These efforts are directed towards enhancing training methodologies, architectural innovations, and ensuring secure implementation (Roberts et al. 2023; Mai et al. 2022). One line of work explored the capabilities of LLMs to translate geospatial tasks expressed in natural language into a set of procedures, and to autonomously interface with external data sources and Geographic Information System (GIS) engines for data retrieval and analysis. Li and Ning (2023) developed a system leveraging the GPT-4 API that autonomously processes spatial problem inputs into a series of operations, retrieves necessary spatial data, generates and performs the coding operations autonomously, and delivers spatial analysis results. Zhang et al. (2023) developed GeoGPT, a framework utilizing GPT-3.5-turbo within the Langchain architecture, to autonomously interpret and execute geospatial tasks by employing appropriate tools from a predefined GIS tool pool.

Further research has examined LLMs' intrinsic geospatial reasoning capabilities. Despite challenges with numerical spatial analysis and abstract reasoning, LLMs have shown proficiency in recalling geographic facts, understanding spatial relationships, and applying this knowledge to practical tasks in real-world contexts that do not require precise spatial calculations. Cohn (2023) discovered that ChatGPT-4 has a correctness range of 67–72 percent in generating RCC-8 composition tables. Ji and Gao (2023) encoded textual descriptions of geometric entities using GPT-2 and BERT, finding that while these LLMs captured certain types of geometry and spatial relationships, they faltered in accurately estimating numeric values and retrieving spatially related objects. Mai et al. (2022) demonstrated that GPT-2 and GPT-3 outperformed fully-supervised, task-specific models in semantic geospatial tasks but struggled with predicting accurate geographic coordinates for recognized toponyms. Roberts et al. (2023) showed that GPT-4 faces challenges in optimizing travel routes based on text-based graph descriptions but excels in recalling detailed geographic information, such as population and life expectancy, and performing spatial reasoning in tasks like generating tourism itineraries. Despite these limitations, we deem that the practical geospatial applications LLMs demonstrate justify continued research into their intrinsic geospatial capabilities, both as a precaution regarding the types of errors they might produce and as guidance for further developmental efforts. This study focuses on the potential for distortions in human cognitive maps to manifest in LLMs, exploring a novel aspect of LLMs' interaction with geographic information.

*Distortions in cognitive maps*

Systematic distortions in the memory of maps are a well-known phenomenon in cognitive psychology, reflecting the ways in which our mental representations of spatial information differ from the actual geography. One notable distortion is referred to as hierarchical organization. It was demonstrated in many studies that human spatial memory tends to be organized hierarchically or categorically, with spatial information clustered into groups such as states or countries (Tversky 1992). When evaluating spatial relationships, such as distance or direction, between points belonging to different groups, the perceived spatial relationship between the larger groups can skew the judgment about the actual relationship between the individual points.

The canonical example of this tendency comes from Stevens and Coupe (1978), who reported on an experiment in which subjects in San Diego, California were asked to indicate from memory the direction to Reno Nevada by drawing a line in the proper orientation on a circle with north noted at the top. Most subjects indicated that Reno is northeast of San Diego, while it is, in fact, northwest. Stevens and Coupe (1978) argued that rather than memorizing the precise locations of every city or the relative positions of all cities, we instead remember the relative locations of states. Within this framework, cities are then categorized and recalled based on the state they are in. Therefore, when people are asked to judge the direction between cities, they do not evaluate it directly but rather, they deduce the relative positions of the cities based on the locations of the states they belong to. In this example, because California is mostly west of Nevada, people often incorrectly assume that all cities in California are west of every city in Nevada. Subsequent experiments demonstrated categorization by states and countries, and by conceptual categories, such as university buildings vs official city buildings vs commercial buildings (Wilton 1979; Maki 1981; Hirtle and Jonides 1985; Chase 1983; Hirtle and Mascolo 1986; Tversky 1992).

*Potential for human-like spatial bias in LLMs*

Recent papers have exposed a range of biases in LLMs (see Gallegos et al. 2023 for a review). These biases include logical fallacies such as the "Reversal Curse"—where models trained on statements like "A is B" struggle to recognize that "B is A" implies the same relationship in reverse (Berglund et al. 2023). Cohn (2023) similarly found that ChatGPT-4 sometimes reasons correctly about a relation but not its inverse. LLMs also include cognitive biases akin to those found in human reasoning, like the anchoring effect, where initial information disproportionately influences subsequent judgments (Jones and Steinhardt 2022). These biases underscore the complex relationship between an LLM's training data, its learning mechanisms, and its output.

While human biases in spatial reasoning are rooted in mental mapping (Tversky 1992), which LLMs do not possess, we hypothesize that they may exhibit similar biases, based on three considerations. Firstly, the biases in human spatial reasoning that emerge from mental mappings might find their way into textual descriptions, which serve as the primary dataset for LLMs. These texts can contain direct inaccuracies influenced by common misconceptions or oversimplified understandings of geography, essentially encoding human biases into data that LLMs learn from. Secondly, the simplification of geographic details in textual narratives could lead LLMs to adopt these generalized views. Just as humans often condense spatial relationships for ease of understanding, the textual information absorbed by LLMs may lack the nuance of actual geographic layouts, encouraging a simplified, and sometimes incorrect, replication of spatial reasoning. Lastly, LLMs might develop biases through their inclination to follow conceptual associations learned from the data (Vaswani et al. 2017) even when they conflict with geographic accuracy.

**Methods**

To answer our research question, we devised a benchmark involving questions on intercardinal directions among well-known American cities and applied it to the models GPT-3.5, GPT-4, and Llama-2.

Our benchmark consists of 14 questions. Among these, seven questions are structured to challenge the LLMs by presenting scenarios where the orientation of larger geographical units could influence the interpretation of directions between cities within them. The remaining seven questions have locations and directions that are less susceptible to the influence of hierarchical categorization, serving as a contrast to the first set. We further varied the pairs to exclude other potential sources of bias. The pairs included various larger geographical units that might induce bias, such as state (in the United States), country, and continent. The locations in nine of the questions have a predominant East-West orientation, while the remaining five questions have a North-South orientation.

We applied the benchmark to Llama-2 70b (Touvron et al. 2023), an open-source model, and GPT-3.5 and GPT-4, developed by OpenAI (2023). Following is an overview of the differences between the models.

- Model size: GPT-4, with its 1 trillion parameters, significantly exceeds GPT-3.5's capacity of 175 billion parameters (Brown et al. 2020) and Llama-2 with 70 billion parameters (Minaee et al. 2024). Models with more parameters are able to assimilate and maintain more information, potentially enhancing their performance across various assessments (Achiam et al. 2023).
- Training Data: GPT-3.5 was trained on a dataset totaling 300 billion tokens and 17 gigabytes, comprising filtered content from Common Crawl, WebText2, and various books and Wikipedia articles (Minaee et al. 2024). GPT-4 utilized a dataset of 13 trillion tokens and 45 gigabytes. Llama-2, introduced in the same year, deployed a training corpus of 2 trillion tokens across its various versions (7B, 13B, 34B, 70B), sourced from diverse online platforms. Models with larger training datasets are able to assimilate and retain more information, likely leading to improved performance across a range of tasks.
- Generalization and Reasoning: In benchmark evaluations focused on generalization and reasoning skills, GPT-4 surpasses both GPT-3.5 and Llama-2 70B's performance (Minaee et al. 2024). It demonstrates superior abilities in logic-based, multi-choice reading comprehension tasks, suggesting an enhanced capacity for complex problem-solving (Achiam et al. 2023). While GPT-3.5 shows inferior performance in various metrics compared to its successor and Llama-2 70B, the latter's smaller parameter size may offer advantages in terms of computational efficiency (Minaee et al. 2024).

We are not familiar with works that compared the capabilities of these specific models in geospatial tasks. However, GPT-2 and BERT were found to have comparable performance in encoding and analyzing geometric entities and spatial relations, a similarity attributed to their analogous tokenization processes and transformer-based architectures (Ji and Gao 2023). In semantic geospatial tasks, Mai et al. (2022) demonstrated that GPT-3 outperforms GPT-2, a difference attributed to the greater number of model parameters.

We asked each model the same set of 14 questions, each posed 10 times. To ensure that each query was treated independently, we reset the model after each question to enable a 'zero-shot' mode, thereby preventing the model from being influenced by previous questions. The format for the questions was consistent: "What is the intercardinal direction from [City 1] to [City 2]?" All questions are presented in Appendix A.

## Results

Performance by question type is summarized in Table 1, and by the question in Table B1 in Appendix B. The first two questions involve directions between cities within the same state. In Question 1, asking about the direction from Dallas to San Antonio, both in Texas, all models achieved perfect scores. Question 2, focusing on the direction from San Antonio to Houston, also in Texas, had GPT 3.5 and GPT 4 maintaining perfect scores, while Llama-2 scored 4/10 (40 percent). This drop in Llama-2's performance might be attributed to the close proximity and slight southern position of San Antonio relative to Houston.

Questions 3-5 involve cities that are situated in directions opposite to the predominant relationships between their respective states. Question 3 is based on the San Diego to Reno scenario from Stevens and Coupe (1978). Question 4 inquiries about the direction between Memphis, Tennessee and Milwaukee, Wisconsin. While Tennessee is mostly east of Wisconsin, Memphis lies west of Milwaukee. In these questions, the models scored only 35/90 (38.9 percent) correct answers. For comparison, question 6, from Minneapolis, Minnesota to Chicago, Illinois, is a simple between-state question since Illinois is completely east of Minnesota, aligning with the city-to-city direction. All models scored perfectly on this question.

Questions 7-11 feature cities in different countries. In 7 and 8, the cities are situated in the opposite direction relative to their category with suspected bias as compared to each other. For instance, in Question 7, Toronto is south of Portland, whereas Canada is north of the United States. This question draws from the findings of Stevens and Coupe (1978), where most respondents incorrectly answered that Toronto is north of Portland, a misconception likely stemming from the general positioning of Canada north of the United States (Figure 1). In Questions 7 and 8, LLMs scored 4/60 (7 percent). Conversely, Questions 9 and 10, where the hierarchical categories (Cuba and Santo Domingo, respectively), do not have a predominant directional relationship with the United States, show alignment in the direction of the cities with their countries. Here, LLMs scored 48/60 (80 percent). Question 11, which compares Ecuador and the United States, with no suspect bias related to the relative position of North and South America, resulted in an LLMs score of 20/30 (67 percent) for the directional comparison between Quito and New York City (NYC). Notably, GPT-4 exhibited exceptionally poor performance in this question, scoring 0/10.

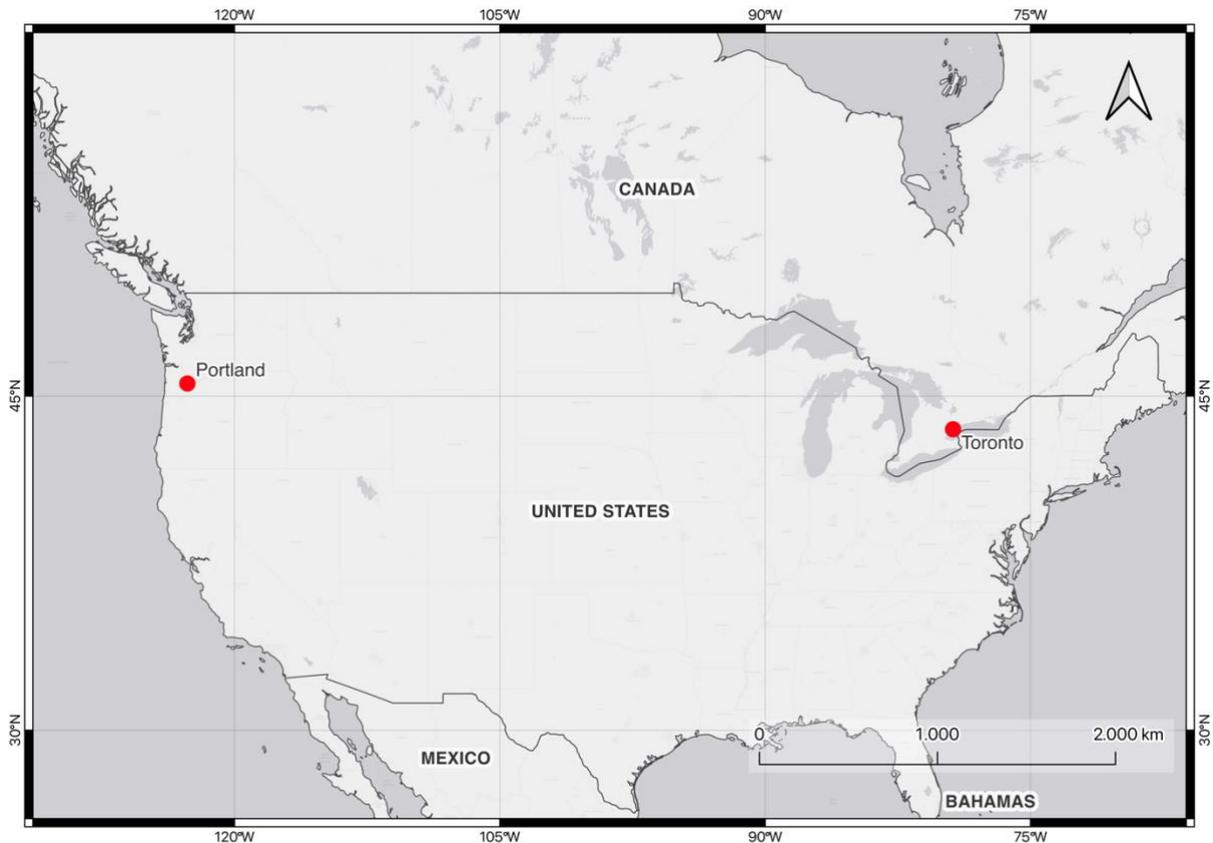

Figure 1. An example representation of hierarchical bias. Although Toronto is geographically south of Portland, it may be perceived as being more north due to its location in Canada, which is predominantly north of the United States, where Portland is situated.

Questions 12-14 in our study are inspired by the concept of alignment, as described by Tversky (1992). This concept refers to the human tendency to perceive objects that are grouped together as being more aligned than they actually are. A notable example is the common perception of North America and Europe as being aligned east-west, when in reality, Europe is largely north of the United States. This leads to a frequent error: locations in southern Europe are often thought to be further south than locations in the northern United States, even when the opposite is true. While a categorization of mapping bias is beyond our scope, we group this phenomenon under hierarchy bias, as it involves a misalignment of objects based on the perceived spatial relationship between their broader geographical areas. Note that the term "alignment" is distinct from the concept of AI alignment, which refers to the process and goal of ensuring that artificial intelligence systems operate in accordance with human values and ethical principles (Gabriel 2020). We revisited Tversky's (1992) experiments, focusing on determining directions between Rome, Italy and Philadelphia, Pennsylvania, and between Monaco and Chicago, Illinois. Additionally, we introduced a straightforward example: the direction from Lisbon, Portugal to NYC, United States, where the common perception accurately aligns with reality—Lisbon is indeed south of NYC. All of the models achieved perfect results (10/10) on the Lisbon-NYC question, while on both the Rome-Philadelphia and Monaco-Chicago questions, all models got all answers wrong (0/20).

Overall, GPT-4 shows the highest accuracy (55.3 percent), followed by GPT-3.5 (47.3 percent) and Llama-2 (44.7 percent). When it comes to discerning directions with suspected hierarchical bias, GPT-4 significantly outperforms the others, scoring 32.9 percent compared to GPT-3.5's 15.7 percent and Llama-2's 7.1 percent. In tasks without suspected hierarchical bias, all models perform notably better, with scores above 85 percent. There is no observable difference in performance between North-South and East-West orientations. While GPT-4 demonstrates superior performance compared to the other models at the in-state and state levels, we cannot identify a distinct pattern in the hierarchical scale category beyond this.

Finally, we also calculated the angular deviations between the cardinal directions (North-South and East-West) and the lines connecting different city pairs. These deviations, shown in Table B1, represent how closely the direction from one city to another aligns with the nearest cardinal direction, varying from 0 to 45 degrees. Generally, as the deviation angles decrease, the accuracy of LLMs decreases. To clarify this concept, consider the Portland to Toronto direction, which is southeast, with a slight 4-degree deviation from the east-west axis. This minimal deviation may increase the challenge of recognizing it as southeast rather than east, illustrating why smaller deviation angles can lead to lower accuracy in identifying the correct intercardinal direction. However, correct answers were still given for questions with low deviation angles, and vice versa. In all of the suspect bias questions except one (Wilmington to Philadelphia), the majority of incorrect responses correctly identified the cardinal direction closest to the actual directional angle. For example, models almost always guess that Toronto is east of Portland.

Table 1. Category-cased performance, by model

|  |  | GPT-3.5 | GPT-4 | Llama-2 |
|---|---|---|---|---|
| **Overall (14)** |  | 47.3% | **55.3%** | 44.7% |
| **Suspect Hier. Bias** | **Yes (7)** | 15.7% | **32.9%** | 7.1% |
|  | **No (7)** | 85.7% | 85.7% | **88.6%** |
| **Hierarchical Scale** | **In-State (2)** | **100.0%** | **100.0%** | 70.0% |
|  | **State (4)** | 45.0% | **80.0%** | 37.5% |
|  | **Country (4)** | 32.5% | **52.5%** | 45.0% |
|  | **Continent (4)** | **50.0%** | 25.0% | **50.0%** |
| **Orientation** | **EW (9)** | 47.8% | **56.7%** | 43.3% |
|  | **NS (5)** | 56.0% | **64.0%** | 56.0% |

**Discussion**

Our benchmark assessment of the spatial reasoning capabilities of LLMs reveals that they excel in deducing straightforward intercardinal directions but struggle with tasks influenced by hierarchical biases. Yet even in their inaccuracies, the models correctly identified the nearest cardinal direction in the majority of cases. This behavior is not indicative of the models functioning as calculators with innate spatial processing capabilities akin to a geographic information system. Instead, their performance reflects a reliance on associative learning from the textual data in their training sets, which includes human-like biases and misconceptions. Indeed, while the models were able to recall accurate geographic coordinates when prompted, they did not seem to utilize this information in calculating spatial relationships in our querying approach.

This is evident in examples like Portland and Toronto, where GPT-4, despite printing the correct coordinates, incorrectly interpreted the directional alignment. GPT-4's superior performance is possibly the result of a larger and more diverse training set, model size and training methods, with the specifics of the latter largely unknown (see Methods section for model comparison). This aligns with the findings by Mai et al. (2022), who attributed GPT-3's enhanced performance in semantic geospatial tasks, compared to GPT-2, to its greater number of model parameters.

Other mapping biases in human cognition could potentially be reflected in LLMs, through the association of locations with other, prominent geographical units. For example, the rotation bias refers to the human tendency to simplify the orientation of geographical elements to align with their 'natural' orientations, as noted by Braine (1978) and Tversky (1992). This bias was evident in a task where participants, when drawing directions between locations in the San Francisco Bay Area, incorrectly perceived Berkeley as being northeast of Stanford, despite Berkeley actually being northwest of Stanford. This illustrates a common tendency to mentally reorient the area's geography to match the north-south axis, despite the Bay Area's actual northwest-southeast diagonal alignment. Similarly, rotation bias might manifest in LLMs through training text data, where people often express the orientation of geographical elements based on their natural orientations. Cities along the United States west coast, for instance, might be contextually interpreted by these models as more westerly positioned compared to inland areas, regardless of their actual coordinates.

Furthermore, our focus on major cities, predominantly within the United States, presents a limitation in the scope of our benchmark's application. This choice was motivated by the availability of well-known geographic locations that are likely to be represented in the training data of the evaluated LLMs. However, the spatial reasoning capabilities of LLMs regarding less-documented settlements, including rural areas and smaller towns, especially in underdeveloped countries, could differ significantly from those observed in this study for several reasons. These places are less likely to be represented in the vast textual data sets LLMs are trained on, potentially leading to a reduction in accuracy and less consistent mistakes than the ones revealed in this paper. Moreover, in the context of the hierarchical spatial bias, the associative learning mechanisms of LLMs may rely on broader, more generalized geographic concepts in the absence of detailed, localized information. To better understand the spatial reasoning capabilities of LLMs, future work should aim to include a more diverse set of locations. This includes not only lesser documented settlements but also places with complex geographic features that challenge the hierarchical categorization tendencies observed in urban American contexts.

What should we do about bias in the spatial memory and reasoning of LLMs and their overall capabilities in intrinsic geospatial analysis? One possibility is to reconsider their utility for such tasks altogether. As Bender et al. (2021) have posited, LLMs can be seen as stochastic parrots, merely echoing the patterns found within their vast training datasets without genuine understanding, and precise logic should not be expected of them. In light of this perspective, efforts could continue in the integration of LLMs with external GIS engines, as explored in recent works by Li and Ning (2023) and Zhang et al. (2023). This certainly makes sense for tasks that involve complex logical reasoning and that do not require high accuracy.

However, given the effectiveness of LLMs in geospatial tasks that prioritize knowledge integration over pinpoint spatial accuracy, and their inherent flexibility, enhancing their geospatial reasoning capabilities could allow solving simple tasks without requiring specialized GIS engines. Given that the bias we discovered likely stems, at least in part, directly from mistakes and simplifications in the description of geographic entities in the training data, high-quality data and detailed descriptions of entities can improve model performance. In addition, the models can be trained directly on spatial relationships of interest, to both improve memory recall and the ability to make inferences about unknown relationships (Ji and Gao 2023). Finally, drawing on the insights from Mai et al. (2022), enhancing the model architecture to incorporate spatial reasoning and align representations of different modalities like geo-tagged texts and remote sensing (or street-view) images could play a crucial role in addressing spatial biases and limited geospatial capabilities.

## Conclusion

We develop and apply a benchmark on three LLMs to assess their ability to discern intercardinal directions. In our findings, the models show distinct patterns in their performance: They fare better in questions without categorization-based distortion in spatial perception, achieving accuracy scores above 85 percent, indicating a strong understanding of straightforward geographical relationships. On the other hand, for questions that were influenced by hierarchical bias—where the spatial relationship could be skewed by broader geographical categories like states or countries—the models displayed much lower accuracy, with GPT-4 achieving only 32.9 percent success in such tasks. In the upcoming phase of our work, we plan to explore additional biases inherent in LLMs' spatial memory, including potential biases related to cultural, linguistic, and contextual factors.

**Author Biographies**

NIR FULMAN is a Post Doctoral Researcher in the GIScience (Geoinformatics) chair in the Department of Geography at Heidelberg University, Heidelberg, Germany. E-mail: nir.fulman@uni-heidelberg.de. His research focuses on simulation and spatial analysis of urban and regional phenomena, urban mobility, and spatial epidemiology.

ABDULKADIR MEMDUHOĞLU is an assistant professor in the Department of Geomatic at Harran University in Şanlıurfa, Türkiye, and a guest researcher in GIScience (Geoinformatics) in the Department of Geography at Heidelberg University, Heidelberg, Germany. Email: memduhoglu@uni-heidelberg.de. His research interests include integrating geospatial data, applying semantic web technologies, and using machine learning to analyze spatiotemporal patterns in GIS and enrich OpenStreetMap (OSM).

ALEXANDER ZIPF is a Professor and Chair of GIScience (Geoinformatics) chair in the Department of Geography at Heidelberg University, Heidelberg, Germany. E-mail: zipf@uni-heidelberg.de. His main research areas include Volunteered Geographic Information (VGI), Crowdsourcing, Citizen Science, and the analysis and processing of geographic data. He focuses on integrating new methods from Geoinformatics and GIScience into Geography.

**Appendix A**

**Questions**

1. What is the intercardinal direction from Dallas, Texas to San Antonio, Texas? Just the direction, nothing else.

2. What is the intercardinal direction from San Antonio, Texas to Houston, Texas? Just the direction, nothing else.

3. What is the intercardinal direction from Wilmington, Delaware to Philadelphia, Pennsylvania? Just the direction, nothing else.

4. What is the intercardinal direction from San Diego, California to Reno, Nevada? Just the direction, nothing else.

5. What is the intercardinal direction from Memphis, Tennessee to Milwaukee, Wisconsin? Just the direction, nothing else.

6. What is the intercardinal direction from Minneapolis, Minnesota to Chicago, Illinois? Just the direction, nothing else.

7. What is the intercardinal direction from Tijuana, Baja California to San Antonio, Texas? Just the direction, nothing else.

8. What is the intercardinal direction from Portland, Oregon to Toronto, Canada? Just the direction, nothing else.

9. What is the intercardinal direction from Santo Domingo, Dominican Republic to Miami, Florida? Just the direction, nothing else.

10. What is the intercardinal direction from Havana, Cuba to Philadelphia, Pennsylvania? Just the direction, nothing else.

11. What is the intercardinal direction from New York City, New York to Quito, Ecuador? Just the direction, nothing else.

12. What is the intercardinal direction from Monaco to Chicago, Illinois? Just the direction, nothing else.

13. What is the intercardinal direction from Rome, Italy to Philadelphia, Pennsylvania? Just the direction, nothing else.

14. What is the intercardinal direction from Lisbon, Portugal to New York City, New York? Just the direction, nothing else.

**Appendix B**

Table B1. Benchmark question details and performance, by model

| No | Cities | GPT-3.5 | GPT-4 | Llama-2 | Suspect Hier. Bias | Hierarchical Scale | Deviation Angle (°) | Orientation | Correct Ans. | Majority Wrong Ans. |
|---|---|---|---|---|---|---|---|---|---|---|
| 1 | Dallas to San Antonio | 10 | 10 | 10 | No | In-State | 23 | NS | Southwest | - |
| 2 | San Antonio to Houston | 10 | 10 | 4 | No | In-State | 9 | EW | Northeast | Southeast |
| 3 | Wilmington to Philadelphia | 0 | 10 | 5 | Yes | State | 35 | EW | Northeast | Northwest |
| 4 | San Diego to Reno | 8 | 2 | 0 | Yes | State | 17 | NS | Northwest | Northeast |
| 5 | Memphis to Milwaukee | 0 | 10 | 0 | Yes | State | 12 | NS | Northeast | Northwest |
| 6 | Minneapolis to Chicago | 10 | 10 | 10 | No | State | 37 | EW | Southeast | - |
| 7 | Tijuana to San Antonio | 3 | 1 | 0 | Yes | Country | 11 | EW | Southeast | Northeast |
| 8 | Portland to Toronto | 0 | 0 | 0 | Yes | Country | 4 | EW | Southeast | Northeast |
| 9 | Santo Domingo to Miami | 10 | 10 | 10 | No | Country | 38 | EW | Northwest | - |
| 10 | Havana to Philadelphia | 0 | 10 | 8 | No | Country | 20 | NS | Northeast | Northwest |
| 11 | New York City to Quito | 10 | 0 | 10 | No | Continent | 6 | NS | Southwest | Southeast |
| 12 | Monaco to Chicago | 0 | 0 | 0 | Yes | Continent | 2 | EW | Southwest | Northwest |
| 13 | Rome to Philadelphia | 0 | 0 | 0 | Yes | Continent | 2 | EW | Southwest | Northwest |
| 14 | Lisbon to New York City | 10 | 10 | 10 | No | Continent | 2 | EW | Northwest | - |
| | **TOTAL** | **47.3%** | **55.3%** | **44.7%** | | | | | | |